\documentclass[11pt,a4paper]{article}
\usepackage[hyperref]{emnlp2020}
\usepackage{times}
\usepackage{latexsym}

\usepackage{microtype}

\aclfinalcopy 

\usepackage{graphicx}
\usepackage{amsmath}
\usepackage{amssymb}
\usepackage{multirow}
\usepackage{makecell}
\usepackage{url}

\usepackage{pgfplots}
\usepackage{booktabs}
\usepackage{subcaption}
\usepackage{graphicx}
\usepackage{verbatim}
\usepackage{url}
\usepackage{array}
\usepackage{balance}
\usepackage{color}
\usepackage{soul}
\usepackage{hyperref}
\usepackage{xcolor}
\usepackage{threeparttable}
\newcolumntype{L}[1]{>{\raggedright\let\newline\\\arraybackslash\hspace{0pt}}m{#1}}
\setul{2pt}{2pt}
\definecolor{skyblue}{RGB}{70, 130, 180}
\usepackage{pifont}
\newcommand{\xmark}{\ding{55}}

\newcommand*{\affaddr}[1]{#1} 
\newcommand*{\affmark}[1][*]{\textsuperscript{#1}}

\title{Aspect Sentiment Classification with Aspect-Specific Opinion Spans}

\author{%
Lu Xu\affmark[* 1, 2]\thanks{$*$ Lu Xu is under the Joint PhD Program between Alibaba and Singapore University of Technology and Design.},
\thanks {Accepted in EMNLP 2020 (Conference on Empirical Methods in Natural Language Processing).}
Lidong Bing\affmark[2], Wei Lu\affmark[1], Fei Huang\affmark[2]\\
\affaddr{\affmark[1]StatNLP Research Group, Singapore University of Technology and Design}\\
\affaddr{\affmark[2]DAMO Academy, Alibaba Group}\\
\tt{xu\_lu@mymail.sutd.edu.sg, luwei@sutd.edu.sg}\\
\tt{\{l.bing, f.huang\}@alibaba-inc.com}\\
}
\renewcommand\footnotemark{}
\date{}

\begin{document}

\maketitle
\begin{abstract}
Aspect sentiment classification, predicting the sentiment polarity of given aspects, has drawn extensive attention. 
Previous attention-based models emphasize using aspect semantics to help extract opinion features for classification. 
However, these works are either not able to capture opinion spans as a whole or capture variable-length opinion spans.
In this paper, we present a neat and effective multiple CRFs based structured attention model that is capable of extracting aspect-specific opinion spans. 
The sentiment polarity of the target is then classified based on the extracted opinion features and contextual information.
The experimental results on four datasets demonstrate the effectiveness of the proposed model, and our further analysis shows that our model can capture aspect-specific opinion spans.\footnote{Our code is released at \url{https://github.com/xuuuluuu/Aspect-Sentiment-Classification}}
\end{abstract}

\section{Introduction}
\label{sec:intro}

Aspect Based Sentiment Analysis (ABSA) \cite{Pang:2008:OMS:1454711.1454712, liu2012sentiment} is an extensively studied sentiment analysis task on a fine-grained semantic level, i.e., opinion targets explicitly mentioned in sentences. Previous ABSA studies focused on a few sub-tasks, such as Aspect Sentiment Classification (ASC) \cite{wang-etal-2016-attention,chen-EtAl:2017:EMNLP20171, ma2018targeted}, Aspect Term Extraction (ATE) \cite{li2018aspect,he-etal-2017-unsupervised},  Aspect and Opinion Co-Extraction~\cite{P13-1172, wang2017coupled,P18-2094,dai2019neural}, E2E-ABSA (a joint task of ASC and ATE) \cite{li2019unified,he-etal-2019-interactive,li-etal-2019-transferable}, Aspect Sentiment Triplet Extraction (ASTE) \cite{peng2020, xu2020}, etc. 
ASC analyzes the sentiment polarity of given aspects/targets in a review. For example, consider the review sentence ``\textit{\textbf{Food} is usually very good, though occasionally I worry about freshness of \textbf{raw vegetables} in side orders.}'' This review mentions two aspects: \textbf{Food} and  \textbf{raw vegetables}, and for ASC, the objective is to  give a positive sentiment on \textbf{Food} and a negative sentiment on \textbf{raw vegetables}.
Most of the previous works \cite{wang-etal-2016-attention, chen-EtAl:2017:EMNLP20171, liu-zhang:2017:EACLshort, Yang2017AttentionBL,  Li2018ExploitingCT, he-etal-2018-effective, li-lu-2019-learning, Hu2019LearningTD} adopt attention mechanism \cite{DBLP:journals/corr/BahdanauCB14} to capture the semantic relatedness among the context words and the aspect, and learn aspect-specific features for sentiment classification. 

However,  it is challenging for attention-based approaches to consider an opinion span as a whole during feature extraction because they are over-reliant on neural models to learn the context-structural information and perform feature extraction over individual hidden representations.
Previous work \cite{bailin-lu:2018:AAAI2018}  engage structured attention networks~\cite{Kim2017StructuredAN}, which extend the previous attention mechanism to incorporate structure dependencies, to model the interaction among context words, and perform soft-selections of word spans.
In particular, they introduce two hand-coded regularizers to constrain the soft-selection process to attend to few short opinion spans. However, such regularizers disturb the structure dependencies, and their method is not capable of emphasizing aspect-specific opinion spans for sentiment classification.

To better capture opinion features for aspect sentiment classification, we propose the MCRF-SA model, which introduces multiple conditional random fields (CRF)~\cite{Lafferty:2001:CRF} to structured attention model.
While exploiting the advantages of structured attention mechanisms, our model avoids the regularizers by the complementarity among multiple CRFs. We also improve the previous position decay function \cite{lixin2018P18-1087, Tang:ACL2019}  to reduce the importance of context words that are further away from the aspect so as to emphasize aspect-specific opinion spans.
Our multi-CRF layer with the effective decay function extracts  aspect-specific features from different representation sub-spaces to overcome the previous limitations. The experimental results on the four datasets demonstrate the effectiveness of our model, and the analysis shows that the behaviors are in alignment with our intuition.

\section{Model Description}

\begin{figure}[!t]
    \centering
    \includegraphics[width=\columnwidth]{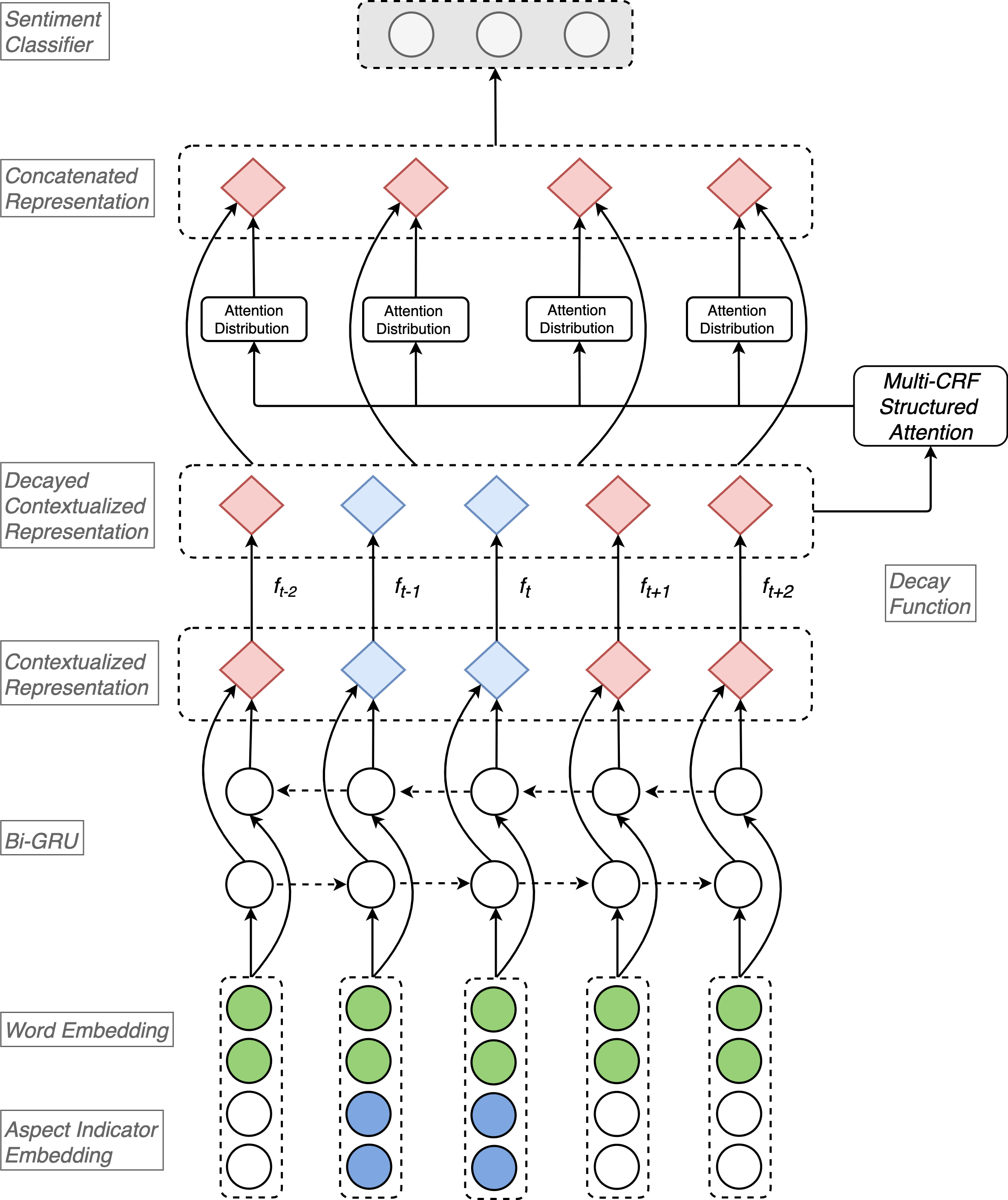}
    \caption{MCRF-SA Architecture.}
    \label{fig:architecture}
    \vspace{-1mm}

\end{figure}

Given a context sequence $\mathbf{w^c} = \{w_1,w_2,\dots, w_n \}$ and a aspect sequence $\mathbf{w^a} = \{w_i, ..., w_j\}$ ($1\leq i\leq j\leq n$) which is a sub-sequence of $\mathbf{w^c}$, the goal of ASC is to predict sentiment polarity $y \in$ \{{\em positive}, {\em negative}, {\em neutral}\} over the given aspect.
Our model is mainly constructed with a few neural layers, including an input layer, an aspect-specific contextualized representation layer, a position decay layer, a multi-CRF structured  attention layer, and a sentiment classification layer. Figure \ref{fig:architecture} presents the architecture of our MCRF-SA model.




\subsection{Input Layer}
The input of our model consists of word embedding $\mathbf{w}^{word}_t$ and aspect indicator embedding $\mathbf{w}^{as}_t$. The aspect indicator embedding is to differentiate aspect words and context words and is randomly initialized.  The input representation $\mathbf{x}_t$ is as follows:
\setlength{\abovedisplayskip}{-5pt} \setlength{\abovedisplayshortskip}{-5pt}
\begin{equation}
\mathrm{\mathbf{x}_t}= [ \mathbf{w}^{word}_t; \; \mathbf{w}^{as}_t]
\end{equation}




\subsection{{Aspect-Specific Contextualized Representation}}
We employ a bi-directional GRU~\cite{Cho2014OnTP} to generate the contextualized  representation.
Since the input representation has already contained the aspect information,  the aspect-specific contextualized representation is obtained by concatenating the hidden states from both directions:
\setlength{\abovedisplayskip}{7pt} \setlength{\abovedisplayshortskip}{7pt}
\setlength{\belowdisplayskip}{7pt} \setlength{\belowdisplayshortskip}{7pt}
\begin{equation}
    \mathrm{\mathbf{h}_t} = [\overrightarrow{\mathbf{h}_t} ; \; \overleftarrow{\mathbf{h}_t}]
\end{equation}
where $\overrightarrow{\mathbf{h}_t}$ is the hidden state from the forward GRU and $\overleftarrow{\mathbf{h}_t}$ is from the backward.
\subsection{Position Decay}
Following the previous work \cite{lixin2018P18-1087, zhang-etal-2019-aspect, Tang:ACL2019}, we also use a position decay function to reduce the influence of the context words on the aspect as it goes further away from the aspect. 
We propose a higher-order decay function, which is more sensitive to distance, and the sensitivity can be tuned by $\gamma$ on different datasets. 
\setlength{\abovedisplayskip}{2pt} \setlength{\abovedisplayshortskip}{2pt}
\setlength{\belowdisplayskip}{8pt} \setlength{\belowdisplayshortskip}{8pt}
\begin{equation}
    f(t)=
    \begin{cases} 
      {(\frac{L \; - \;i\; +\; t} {L})}^{\gamma}  & t < i \\
      1 & i\leq t\leq j\\
      {(\frac{L\; -\; t\; +\;  j}{L})}^{\gamma}  & j < t \\
    \end{cases}
\end{equation}
where $i$ and $j$ are the starting and ending position of an aspect, $L$ is the maximum length of sentences across all datasets, $\gamma$ is a hyper-parameter and a larger value enables more influence from the context words that are close to the aspect.
Then, the decayed contextual word representation is as follows:
\begin{equation}
    \mathbf{r}_t = f(t) \; \mathbf{h}_{t}
\end{equation}

\subsection{Multi-CRF Structured Attention}
We use multiple linear-chain CRFs to intensively incorporate structure dependencies to capture the corresponding opinion spans of an aspect. In particular, we create a latent label \cite{bailin-lu:2018:AAAI2018} $z \in \{Yes, \; No\} $ to indicate whether each context word belongs to part of opinion spans.  Similar to \cite{lample-etal-2016-neural}, given the sentence representation $\mathbf{x}$, the CRF is defined as:
\setlength{\abovedisplayskip}{6pt} \setlength{\abovedisplayshortskip}{6pt}
\setlength{\belowdisplayskip}{6pt} \setlength{\belowdisplayshortskip}{6pt}
\begin{equation}
    P(\mathbf{z} | \mathbf{x}) = \frac{\exp(score( \mathbf{z},  \mathbf{x} ))}{\sum_{\mathbf{z}^{'}} \exp(score( \mathbf{z}^{'}, \mathbf{x}))}
    \label{eq:prob}
\end{equation}
\textcolor{black}{where $score(\mathbf{z},\mathbf{x})$ is a score function that is defined as the summation of transition scores and emission scores from the Bi-GRU:}
\begin{equation}
    score(\mathbf{z}, \mathbf{x}) = \sum_{t=0}^{n} T_{z_t, z_{t+1}} + \sum_{t=1}^{n} E_{\mathbf{t}, z_t }
\end{equation}
where $T$ is a transition matrix and ${T_{z_{t}, z_{t+1}}} $ denotes the transition score from label $z_t$ to $z_{t+1}$.
 $E_{\mathbf{t}, z_t }$ denotes the emission score of label $z_t$ at the $t$-th position, and the score is obtained from a linear layer, which takes  $ \mathbf{r}_t$ as input and returns a vector whose length is label size.


\subsubsection{Marginal Inference}
The latent labels introduced in the CRF layer show whether the word influences the given aspect's sentiment.  \textcolor{black}{Intuitively, we can understand that the marginal probabilities on the $Yes$ label indicate the influence of the current context word on the aspect word's sentiment. }
 By using the forward-backward algorithm, we calculate the marginal distribution of the latent label. With the marginal distribution, the sentence representation $\mathbf{s}$ is obtained:
\begin{equation}
\label{head_rep}
\mathbf{s} = \sum_{t=1}^{n} P(z_t = Yes | \mathbf{x}) \mathbf{r}_t
\end{equation}
 The final  representation for classification is obtained by concatenating the sentence representations from all CRFs:
\begin{equation}
    \mathbf{q} = [\mathbf{s}_1; \mathbf{s}_2; ...; \mathbf{s}_a]
    \label{eq:sent_rep}
\end{equation}
where $a$ is the number of CRFs.

\subsection{Sentiment Classification}

The sentence representation $\mathbf{q}$ is passed to a sentiment classier to obtain the distribution of sentiment polarities:
\begin{equation}
    \label{classifier}
    P(\mathbf{y} | \mathbf{q}) =  \mathrm{Softmax}(W\mathbf{q}  + \mathbf{b})
\end{equation}
where $W $ and $\mathbf{b}$ are learnable parameters for the sentiment classifier layer. We learn model parameters by minimizing the negative log-likelihood.


\section{Experiments}
\subsection{Experimental Setup}
Our proposed MCRF-SA model  is evaluated on four benchmark datasets: SemEval 2014 Task4 \cite{pontiki-EtAl:2014:SemEval}, SemEval 2015 Task12 \cite{pontiki-etal-2015-semeval} and SemEval 2016 Task 5 \cite{pontiki-etal-2016-semeval}. 
Following the previous works \cite{Tang2016AspectLS, chen-EtAl:2017:EMNLP20171, bailin-lu:2018:AAAI2018, he-etal-2018-effective}, we remove a few examples that have conflicting labels. Detailed statistics of the datasets can be found in Table \ref{tab:statistics}. 

We use the 300d GloVe \cite{pennington2014glove} to initialize our word embeddings.  One-sixth of instances are randomly selected from the original training dataset as the development dataset, and the model is only trained with the remaining data. With the development set, we tune our model hyper-parameters using an open-source black-box tuner \cite{alberto}. We set the hidden size of GRU to 32 or 64. The batch size is set to 64 or 96. The dropout rate is selected from 0.3 to 0.8, with a step size of 0.1. The dimension of the aspect indicator is selected from \{50, 70, 90\}. The value of $\gamma$ in the position decay function is selected from \{1,2,3\}. The number of layer of GRU is selected from \{1,2,3\}. We adopt Adam~\cite{kingma2014adam} to optimize our model with a learning rate of 0.008. All hyper-parameters are selected based on the best performance on the development set.
\begin{table}[t!]
\centering
\resizebox{\columnwidth}{!}
{%
\begin{tabular}{@{~}l@{~}c@{~}c@{~}c@{~}c@{~}c@{~}c@{~}c@{~}c@{~}c@{~}}
\toprule
 \multirow{2}{*}{\textbf{Dataset}}  & \multicolumn{3}{c}{\textbf{Train}}  & \multicolumn{3}{c}{\textbf{Dev}}  & \multicolumn{3}{c}{\textbf{Test}} \\ \cmidrule(lr){2-4}\cmidrule(lr){5-7} \cmidrule(lr){8-10}
& \#Pos. & \#Neu. & \#Neg. &  \#Pos. & \#Neu. & \#Neg. & \#Pos. & \#Neu. & \#Neg.  \\\hline 
\textbf{14Rest}  & 1796 & 539 & 666  & 368 & 94 & 139 & 728 & 196 & 196 \\
\textbf{14Lap}  &  824 & 383 & 717 &  161 & 72  & 149 & 340 & 167 & 128 \\
\textbf{15Rest} &  808 & 29 & 228 &  147 &5 & 44 &  340 & 28 & 195 \\
\textbf{16Rest} & 1106 & 54 & 406 &  191 & 9 & 60 & 474 & 29 & 127 \\


\bottomrule
\end{tabular}}
\caption{Statistics of datasets.  }
\label{tab:statistics}
\vspace{-2mm}
\end{table}

\subsection{Baselines} 
Our MCRF-SA model is compared with the following methods\footnote{Note that our focus is not on exploring the power of pre-trained models (e.g.,  BERT and ELMo) for ASC.}.
 \begin{table*}[!t]

    \centering
    \resizebox{0.99\textwidth}{!}{%
    \begin{tabular}{llcccccccc}
    \toprule
      & \multirow{2}{*}{\textbf{Models}} & \multicolumn{2}{c}{14Rest} & \multicolumn{2}{c}{14Lap} & \multicolumn{2}{c}{15Rest}  & \multicolumn{2}{c}{16Rest}  \\ \cmidrule(lr){3-4} \cmidrule(lr){5-6}\cmidrule(lr){7-8} \cmidrule(lr){9-10}
       & & $Acc.$ & $F_1$ & $Acc.$ & $F_1$ & $Acc.$ & $F_1$ & $Acc.$ & $F_1$\\  \hline
       \multirow{9}{*}{\textbf{Baselines}} 
        & SVM \cite{kiritchenko-EtAl:2014:SemEval} & 80.16$^\natural$ & - & 70.49$^\natural$ & - & - & - & - & - \\
        &ATAE-LSTM \cite{wang-etal-2016-attention} & 77.20$^\natural$ & - & 68.70$^\natural$ & - & - & -  & - & -\\ 
        &MemNet \cite{Tang2016AspectLS} & 79.61$^*$ & 69.64$^*$ & 70.64$^*$ & 65.17$^*$ & 77.31$^*$ & 58.28$^*$ & 85.44$^*$ & 65.99$^*$  \\
        &IAN \cite{Ma2017InteractiveAN} & 79.26$^*$ & 70.09$^*$ & 72.05$^*$ & 67.38$^*$ & 78.54$^*$ & 52.65$^*$ & 84.74$^*$ & 55.21$^*$  \\        
        & SA-LSTM-P \cite{bailin-lu:2018:AAAI2018} & 81.60$^\natural$ & - & 75.10$^\natural$ & - & - & - & 88.70$^\natural$ & - \\ 
        & TNet-LF \cite{lixin2018P18-1087} & 80.42$^*$ & 71.03$^*$ & 74.61$^*$ & 70.14$^*$ & 78.47$^*$ & 59.47$^*$ & 89.07$^*$ & 70.43$^*$\\
        & TNet-ATT \cite{Tang:ACL2019} & 81.53$^\natural$ & 72.90$^\natural$ & 77.62$^\natural$ & 73.84$^\natural$ & - & - & - & -\\ 
        & ASCNN \cite{zhang-etal-2019-aspect} & 81.73$^*$ & 73.10$^*$ & 72.62$^*$ & 66.72$^*$ & 78.48$^*$ & 58.90$^*$ & 87.39$^*$ & 64.56$^*$\\ 
        & ASGCN \cite{zhang-etal-2019-aspect} & 80.86$^*$ & 72.19$^*$ & 74.14$^*$ & 69.24$^*$ & 79.34$^*$ & 60.78$^*$ & 88.69$^*$ & 66.64$^*$\\\hline
        \multirow{2}{*}{\textbf{Reproduce}\footnote{}} 
        &TNet-ATT \cite{Tang:ACL2019}&79.38{\color{white}$^*$} &69.44{\color{white}$^*$}&76.22{\color{white}$^*$}&71.51{\color{white}$^*$}&-&-&-&-\\
        &ASGCN \cite{zhang-etal-2019-aspect} & 79.73{\color{white}$^*$}& 70.48{\color{white}$^*$} & 72.91{\color{white}$^*$} & 68.06{\color{white}$^*$} & 78.74{\color{white}$^*$}& 57.67{\color{white}$^*$}& 87.71{\color{white}$^*$}& 70.29{\color{white}$^*$}\\ \hline
        {\textbf{Ours}}& MCRF-SA & \textbf{82.86$^{\dag}$} & \textbf{73.78$^{\dag}$} & \textbf{77.64$^{\dag}$} & \textbf{74.23$^{\dag}$} & \textbf{80.82$^{\dag}$} & \textbf{61.59$^{\dag}$} & \textbf{89.51$^{\dag}$} & \textbf{75.92$^{\dag}$}\\ 
    \bottomrule
    \end{tabular}}
    \caption{Experimental results (\%). The results with symbol``$\natural$'' are retrieved from the original papers, and those with $*$ are retrieved from \citet{zhang-etal-2019-aspect}. The marker $^{\dag}$ refers to $p$-value $<$ 0.01 when comparing with ASGCN.}
    \label{tab:main_results}
\end{table*}
SVM~\cite{kiritchenko-EtAl:2014:SemEval} is a support vector machine based method that integrates surface, lexicon, and parse features.
ATAE-LSTM~\cite{wang-etal-2016-attention} is an LSTM \cite{lstm} based model, which has an extra attention to perform soft-selection over the context words.
MemNet~\cite{Tang2016AspectLS} introduces a deep memory network to implement attention mechanisms to learn the relatedness of context words towards the aspect.
IAN~\cite{Ma2017InteractiveAN}  utilizes two LSTM based attention models to learn both context and aspect representations interactively.
SA-LSTM-P~\cite{bailin-lu:2018:AAAI2018} employs structured attention networks with multiple regularizers to capture the opinion spans for ASC. 
TNets~\cite{lixin2018P18-1087} implements a context-preserving mechanism to get the aspect-specific word representations and uses a Convolutional Neural Network (CNN)~\cite{cnn} layer to obtain the sentence representation.
TNet-ATT~\cite{Tang:ACL2019} is an extension of TNet-LF, and it provides an attention supervision mining mechanism to improve the previous model. 
ASCNN and ASGCN \cite{zhang-etal-2019-aspect} use CNN and Graph Convolutional Network (GCN)~\cite{kipf2017semi} to capture the long-range dependencies and syntactic information.

\subsection{Experimental Results}
Our proposed model shows significant improvements on the four datasets, Table \ref{tab:main_results} shows the performance comparisons. 
\footnotetext{We train their models with the default parameters and their released training data, and report the average results on our test sets from 3 runs. Note that these works did not release development sets.}
Our method outperforms SVM~\cite{kiritchenko-EtAl:2014:SemEval} by 2.7 and 7.15 $Acc.$ score on 14Rest and 14Lap, respectively. 
This indicates that our neural approach extracts more effective features than hard-coded feature engineering. 
Compared to the attention-based methods -- ATAE~\cite{wang-etal-2016-attention}, MemNet~\cite{Tang2016AspectLS}, IAN~\cite{Ma2017InteractiveAN}, and TNet-ATT~\cite{Tang:ACL2019},  our MCRF-SA model pays more attention to the aspect-specific opinion spans, which bring significant performance improvement on the four datasets.

We also compare our model with methods that focus on word segmentations for sentiment classification. 
Our method outperforms the previous regularizers guided structured attention model SA-LSTM-P \cite{bailin-lu:2018:AAAI2018} by more than 1.2 $Acc.$ score on 14Rest and 14Lap. 
TNet-LF~\cite{lixin2018P18-1087} and ASCNN \cite{zhang-etal-2019-aspect} employ CNN to evaluate word spans regarding how much it contributed to the sentiment, but the kernel size limits the length of the span.
ASGCN \cite{zhang-etal-2019-aspect} employs GCN over the dependency tree to capture syntactic and dependency information. However, the performance heavily relies on the accuracy of the dependency trees.
Our proposed multi-CRF structured attention along with the position decay function allows MCRF-SA to perform soft-selection of multiple aspect-specific opinion spans that influence the aspect's sentiment. The large performance gaps between our model and baseline models confirm the effectiveness of our proposed architecture. Such results also demonstrate that sentiment classification can benefit greatly from aspect-specific opinion spans.

Furthermore, we observe that the performance on 15Rest is not as good as the other three datasets. Such behavior is caused by the different distribution of positive, neutral, and negative sentiment between training and test set, shown in Table \ref{tab:statistics}.

\section{Analysis}
\subsection{Effect of Number of CRFs}
\begin{figure}
\centering
\resizebox{\linewidth}{!}{
\begin{tikzpicture}
\pgfplotsset{width=7.5cm, height = 4.5cm , compat=1.9}
\begin{axis}
[
            title={},
            xlabel={Number of CRF},
            legend style={font=\fontsize{6}{1}\selectfont},
            ylabel={Accuracy [\%]},
            xmin=0, xmax=17,
            ymin=70, ymax=88,
            label style={font=\fontsize{8}{1}\selectfont},
            xticklabel style = {font=\fontsize{8}{1}\selectfont},
            yticklabel style = {font=\fontsize{8}{1}\selectfont},
            xtick={0,2,4,6,8,10,12,14,16},
            ytick={70.0,74,78,82.0,86.0},
            legend pos = north east,
        ]
\addplot [mark=triangle*, color=orange] plot coordinates {
(1,79.38)(2,79.29)(3,81.70)(4,82.86)(5,81.61)(6,80.36)(8,80.18)(12,81.42)(16,80.36) };
\addplot [mark=o, color= skyblue] plot coordinates {
(1, 74.80) (2, 74.02)(3,76.70)(4, 77.64) (5, 77)(6,76.85)(8,74.65)(12, 74.49)(16, 74.65)
};

\legend{14Rest\\14Lap\\}
\end{axis}
\end{tikzpicture}
}
\caption{{Effect of number of CRFs.}}
\label{fig:heads}
\vspace{-1mm}
\end{figure}
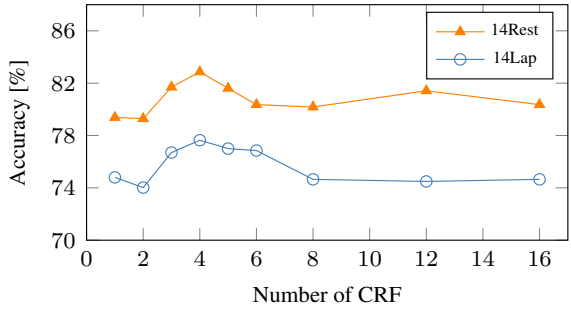

To fully investigate the effect of the number of CRFs, we conduct additional experiments on 14Rest and 14Lap with the number of CRFs $\in \{1,2,3, ..., 16\}$. Figure \ref{fig:heads} shows the experimental results.
The model achieves the best performance when the number of CRFs equals to 4. Particularly, the performance becomes relatively plateau when a large number of CRFs is adopted. We believe this is because the sizes of the four benchmark datasets are relatively small, and an excessively large number of parameters may not be able to further extract effective features. 

\begin{figure}[!t]
\centering
    \begin{subfigure}[b]{\columnwidth}
      \includegraphics[width=1\linewidth]{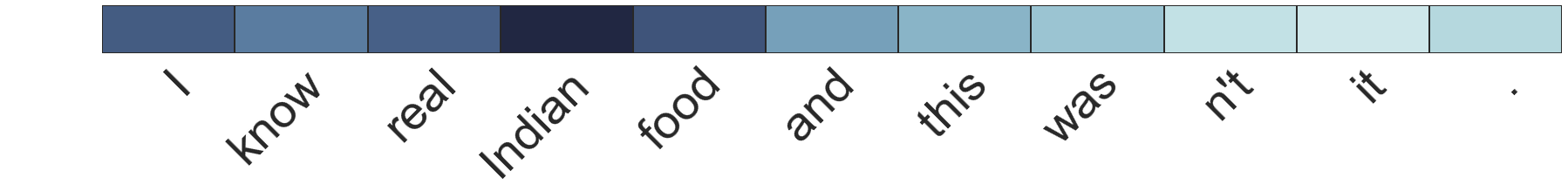}
      \caption{\mbox{SA-LSTM-P.}\label{fig:att2}}
      \vspace{3mm}
    \end{subfigure}
    \begin{subfigure}[b]{\columnwidth}
      \includegraphics[width=1\linewidth]{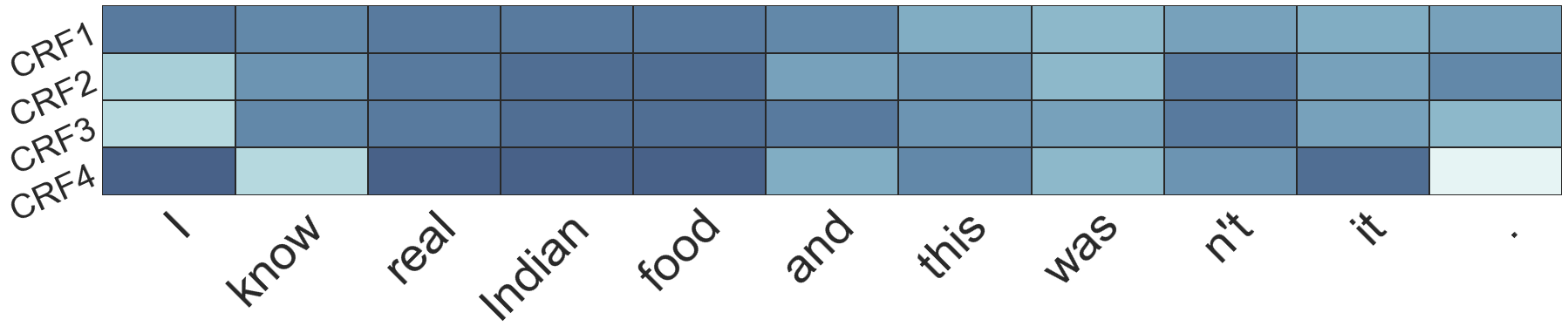}
      \caption{MCRF-SA with 4 CRFs\label{fig:att3}.}
       
    \end{subfigure}

\caption[Attention Scores]{Marginal distributions of "Yes" label.}
\label{fig:case}
\vspace{-1mm}
\end{figure}

\subsection{Case Study and Error Analysis}



Figure \ref{fig:case} shows the marginal distributions (Equation \ref{eq:prob}) of  SA-LSTM-P \cite{bailin-lu:2018:AAAI2018} and our MCRF-SA model. The aspect for the given example is ``\textit{Indian food}'' with negative sentiment, and only our model predicts correct sentiment. 
From Figure \ref{fig:att3} heat map, the different marginal distributions on the four CRFs indicate that our model indeed captures different opinion features. 
It can be observed that  MCRF-SA is able to attend to the two major opinion spans: \textit{``real"} and \textit{``n't"}. 
The SA-LSTM-P model returns positive sentiment as it focuses too much on wrong opinion words. 

We also analyze some common errors from our MCRF-SA model, ASGCN, and TNet-ATT on the Lap14 dataset. We observe two major types of errors, and Table \ref{tab:case_study} shows the examples for error analysis. The first two sentences belong to the type 1 error and the last one presents a type 2 error.
{\color{black}The first type of errors appear frequently in neutral cases. In general, the neural models cannot well differentiate if the negative expressions (e.g. ``{\em cost}'', ``{\em shouldn't}'', etc.) is associated with the target/aspect.} The second type typically involve complicated sentence structures with non-trivial semantics, which requires advanced language understanding capability. 

\begin{table}[!t]
    \centering
    \resizebox{\columnwidth}{!}
    {%
    \Large
    \begin{tabular}{@{}L{9cm}@{~}|@{~}ccc@{}}
    \toprule
        \textbf{Case Study} & {MCRF-SA} & {ASGCN} & {TNet-ATT} \\ \hline
        1. When considering a Mac, look at the total cost of ownership and not just the initial \textcolor{skyblue}{\textbf{price tag}}\textbf{$_\text{NEU}$} .& NEU & NEG$_{\textcolor{orange}{\text{\xmark}}}$  & NEG$_{\textcolor{orange}{\text{\xmark}}}$ \\ \hline
        2. It shouldn't happen like that, I don't have any \textcolor{skyblue}{\textbf{design app}}\textbf{$_\text{NEU}$} open or anything .& NEG$_{\textcolor{orange}{\text{\xmark}}}$   & NEU & NEU\\ \hline
        3. The smaller \textcolor{skyblue}{\textbf{size}}\textbf{$_\text{POS}$} was a bonus because of space restrictions. & NEG$_{\textcolor{orange}{\text{\xmark}}}$  & NEG$_{\textcolor{orange}{\text{\xmark}}}$  & NEG$_{\textcolor{orange}{\text{\xmark}}}$  \\ 
    \bottomrule 
    \end{tabular}}
        \caption{The words highlighted in blue denote the given aspects, and gold sentiment labels are marked as subscripts. $_{\textcolor{orange}{\text{\xmark}}}$  indicates incorrect prediction.}
    \label{tab:case_study}
\end{table}

\subsection{Ablation Study} 
We examine the effectiveness of the major components of our MCRF-SA model, and 
Table 3 presents the ablation results on 14Rest and 14Lap datasets.
Without the aspect indicator, our model becomes a sentence-level sentiment classification method which inevitably produces wrong predictions for sentences having multiple aspects with different sentiments. 
Removing the position decay function hurts the performance by 2.84 and 1.11 $F_1$ score on 14Rest and 14Lap,  respectively. 
Lastly,  without multi-CRF structured attention layer, the architecture becomes a simple Bi-GRU based model and the performance drops significantly by 4.89 and 10.49 $F_1$ points on 14Rest and 14Lap. 
\begin{table}[t!]
\centering
\resizebox{\columnwidth}{!}
{%
    \begin{tabular}{llcccc}
    \toprule
      & \multirow{2}{*}{\textbf{Models}} & \multicolumn{2}{c}{14Rest}& \multicolumn{2}{c}{14Lap}  \\ \cmidrule(lr){3-4} \cmidrule(lr){5-6}
       & & $Acc.$ & $F_1$& $Acc.$ & $F_1$\\ \hline 
        &MCRF-SA & 82.86 & 73.78 & 77.64 & 74.23\\ 
        &~~~~--  aspect indicator & 79.02 & 66.96  & 72.76 & 67.56\\
        &~~~~--  decay function & 81.52 & 70.94 & 76.69 & 73.12\\
        &~~~~--  structured attention & 80.00 & 68.89 & 69.61 & 63.74 \\ 
        
    \bottomrule
    \end{tabular}}
    \caption{Ablation Study. } 

\end{table}

\section{Conclusion}
We propose a simple and effective MCRF-SA model to extract aspect-specific opinion span features. In particular, with the proposed multi-CRF structured attention layer and the effective position decay function, our model is capable of extracting various aspect-specific opinion span features from different representation sub-spaces. The experimental results demonstrate that our method effectively exploits the corresponding opinion features for sentiment classification. One future direction is to investigate how to integrate the two different attention mechanisms, namely the standard attention  and  structured  attention for NLP applications. 

\section*{Acknowledgements}
We would like to thank the anonymous reviewers for their helpful comments.
This research is partially supported by Ministry of Education, Singapore, under its Academic Research Fund (AcRF) Tier 2 Programme (MOE AcRF Tier 2 Award No: MOE2017-T2-1-156). 
Any opinions, findings and conclusions or recommendations expressed in this material are those of the authors and do not reflect the views of the Ministry of Education, Singapore.

\bibliography{emnlp2020}
\bibliographystyle{acl_natbib}

\end{document}